\documentclass[letterpaper, 10 pt, conference]{ieeeconf}  %
\usepackage{amsmath,amssymb}
\usepackage{graphicx}
\usepackage{cite}
\usepackage{algorithm}
\usepackage{algorithmic}
\usepackage{float}      %
\usepackage{tikz}       %
\usetikzlibrary{positioning,calc,fit,backgrounds}
\usepackage{comment}
\usepackage{booktabs}
\usepackage{subcaption}
\usepackage{multirow}
\usepackage{tabularx}
\usepackage{array}
\usepackage[table]{xcolor}
\usepackage{siunitx}           
\usepackage[hidelinks]{hyperref} 
\usepackage{placeins} 
\usetikzlibrary{arrows.meta,positioning,fit,calc,backgrounds,shapes.misc}

\IEEEoverridecommandlockouts                              %

\title{\LARGE \bf
SSQA: Sibling-Selective Quadtree Attention for Hierarchical Modeling in Perception Tasks
}

\author{
Yufan Chen$^{1}$*, 
Arnav Bali$^{2}$*, 
Angela Liu$^{3}$, 
Laura Zheng$^{3}$, 
and Ming C. Lin$^{3}$%
\thanks{Emails: \{kchen503\}@usc.edu, arnavbali36@gmail.com, angelaliu9805@gmail.com, \{lyzheng, lin\}@umd.edu}%
\thanks{*These authors contributed equally to this work.}%
\thanks{*This work was supported in part by the NSF REU-CAAR Program.}%
\thanks{$^{1}$Thomas Lord Department of Computer Science, University of Southern California, Los Angeles, CA, USA.}%
\thanks{$^{2}$South River High School, Edgewater, MD, USA.}%
\thanks{$^{3}$Department of Computer Science, University of Maryland, College Park, MD, USA.}%
}
\begin{document}

\maketitle
\begin{tikzpicture}[remember picture,overlay]
\node[anchor=north,yshift=-0.35in,text width=6.8in,align=center,font=\scriptsize] at (current page.north)
{\copyright~2026 IEEE. Personal use of this material is permitted. Permission from IEEE must be obtained for all other uses, in any current or future media, including reprinting/republishing this material for advertising or promotional purposes, creating new collective works, for resale or redistribution to servers or lists, or reuse of any copyrighted component of this work in other works. Published in: Y.~Chen, A.~Bali, A.~Liu, L.~Zheng, and M.~C.~Lin, ``SSQA: Sibling-Selective Quadtree Attention for Hierarchical Modeling in Perception Tasks,'' \textit{2026 IEEE International Conference on Robotics and Automation (ICRA)}, Vienna, Austria, 2026. DOI: \href{https://doi.org/10.1109/ICRA57385.2026.11697594}{10.1109/ICRA57385.2026.11697594}};
\end{tikzpicture}
\thispagestyle{empty}
\pagestyle{empty}

\begin{abstract}
Perception tasks for navigation in robotics, including aerial platforms such as drones and autonomous driving systems, are inherently structured. Drone-mounted cameras typically capture sky above, terrain below, and obstacles or man-made structures in between, while driving data often contains organized road layouts, lane markings, and surrounding agents. Motivated by these axis-aligned structural priors, we note that such information is typically more structured than in generic image tasks. We hypothesize that processing information in a quadtree-esque manner can not only model features effectively in a hierarchical manner, but also offers an efficient linear-time alternative to vanilla attention mechanisms, which run in quadratic time. In this paper, we propose Sibling-Selective Quadtree Attention (SSQA), which models image tokens hierarchically as a structured, full quadtree. We show analytical complexity analysis that guarantees linear-time feature modeling, in addition to empirical experiments comparing inference speeds with other popular modeling approaches, such as Mamba 2 and Quadtree Attention. Our results, benchmarked across several tasks, show that we achieve results at least as good, if not notably better, as others at a fraction of the computational costs.

\end{abstract}

\section{Introduction}

Perception tasks in computer vision face numerous sources of noise during deployment in autonomous driving and robotics under adverse weather, low-light conditions, and frequent occlusions. While state-of-the-art perception and end-to-end models perform well in the most common, normal settings, given large datasets and a high number of parameters, the capability of such systems to scale to unseen domains remains a challenge -- both in compute feasibility and in performance. 

Motivated by the challenge of scaling deep learning for autonomous perception, we observe that camera and remote sensing data from aerial navigation is \textit{typically more structured than generic image tasks}. For example, as in a remote sensing dataset, aerial imagery often contains sky regions near the top, terrain near the bottom, and man-made structures or vegetation in between. This inherent structure can be leveraged during training as a prior to improve convergence and generalization, particularly for drone perception and navigation. 

Currently, state-of-the-art perception models apply variations of hierarchical attention to encode information at both low and high granularity levels~\cite{liu2021swin,xie2021segformer}. 
Specifically, alternatives such as Linear Transformers~\cite{katharopoulos_et_al_2020_linear_transformers} and Mamba~\cite{mamba, mamba2} have been investigated as efficient linear alternatives to the vanilla Transformer. However, such approaches often trade modeling capacity for efficiency, and may not yield practical speedups without specialized optimization.

In this paper, we propose a compute efficient, performance-preserving attention mechanism based on quadtree organization of information density within the scene, inspired by the structured nature of navigation and remote sensing data. 
The key difference that distinguishes our work from previous efforts in efficient transformer alternatives is the observation that optimizations in efficiency stem from priors. Our hypothesis is that \textit{some pairwise token comparisons are more useful than others}, and that most task features are axis-aligned and can thus be processed hierarchically as a structured tree.

\begin{figure}[t!]
    \centering
    \includegraphics[width=0.9\linewidth]{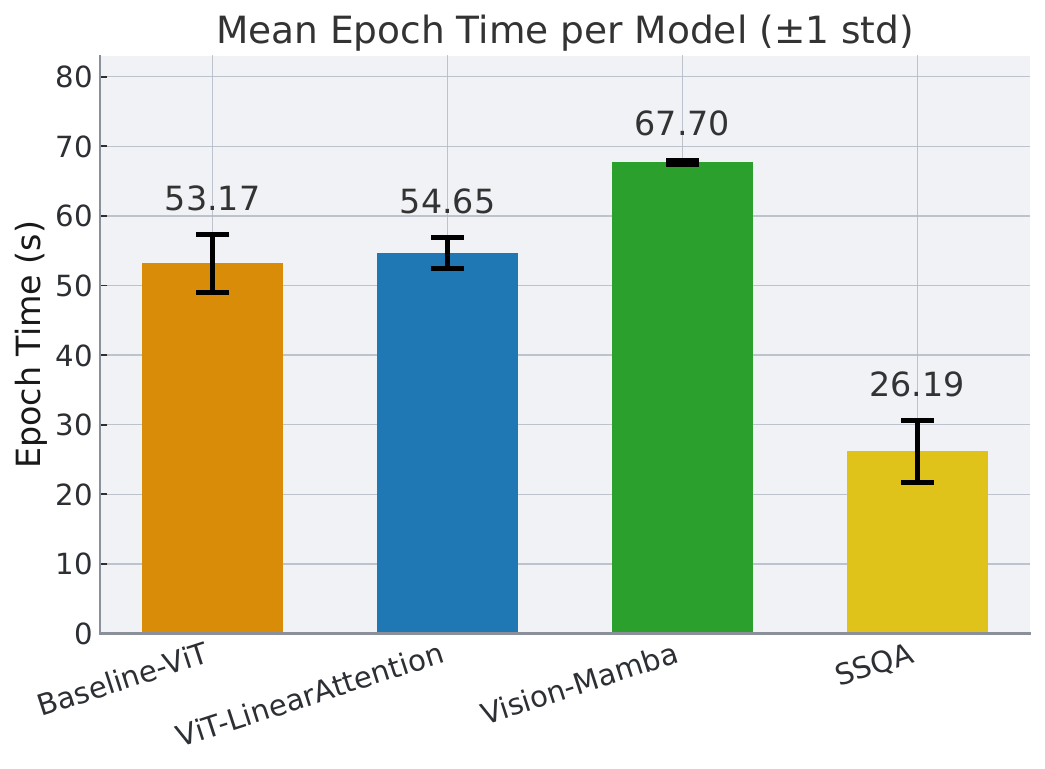}
    \caption{\textbf{Mean Epoch Wall Times (s) across 50 epochs on IllusionAnimals classification (224$^2$ inputs) on A100 GPU.}
    Despite other methods also modeling attention in linear time, a practical caveat is the lack of compute optimization in practice. As a result, some methods are even slower than vanilla ViT~\cite{dosovitskiy2020vit} out-of-the-box from official implementations. SSQA is both easy to integrate and requires no additional practical optimizations, our results show that SSQA maintains or even improves performance over other hierarchical modeling baselines.}
    \label{fig:mean_epoch_time}
\end{figure}

In aerial robotics, perception emphasizes large-scale scene understanding, where drone-mounted cameras typically capture sky regions above, terrain below, and obstacles or man-made structures in between. In autonomous driving, systems must interpret and combine structured cues such as road layout, lane markings and surrounding moving agents to ensure safe operation. Our Sibling-Selective Quadtree Attention (SSQA) is designed to exploit this kind of inherent spatial organization, enabling efficient and robust representation learning across diverse tasks.

\begin{figure*}[t!]
    \centering
    \includegraphics[width=0.78\linewidth]{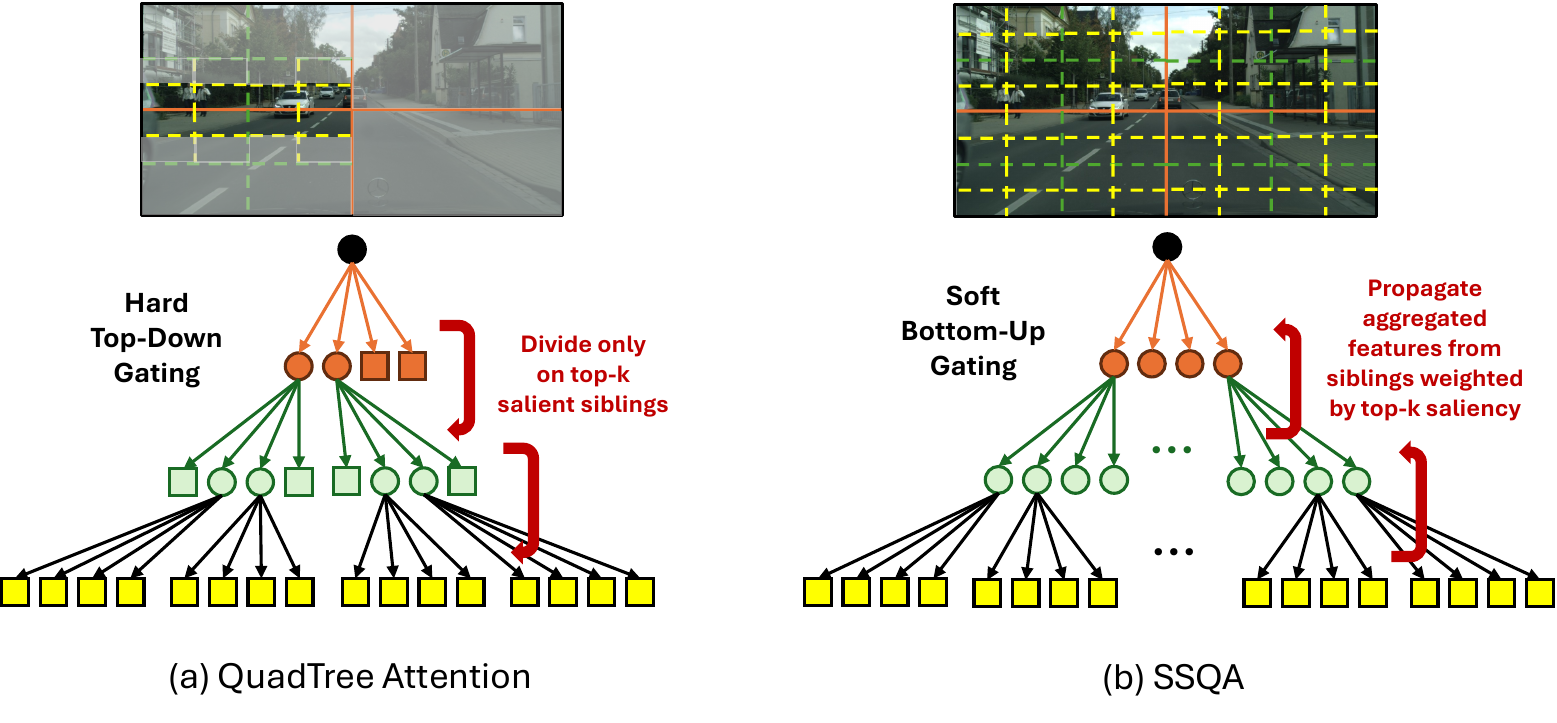}
    \vspace*{-0.25em}
    \caption{\textbf{Top-Down vs. Bottom-Up Gating; Squares indicate leaf nodes.} Our work has a key difference from previous work in hierarchical quadtree-based attention approaches: bottom-up gating. Previous works (a) avoid subdivisions into tokens with smaller granularity if possible by selecting top-k tokens with highest saliency during training. As a result, \textit{tokens on the same level outside of top-k saliency neglect hierarchical reasoning, even if there agents present in those patches}. For example, in a driving scene, one patch may have a higher density of pedestrians, but it shouldn't imply less attention on the single agent falling outside of the same patch. In our approach, we model features from bottom-up, \textit{ensuring that features are modeled hierarchically across the entire image}. Doing so not only improves performance on various benchmarked tasks, but also maintains linear complexity with respect to number of tokens (shown in \S\ref{sec:complexity}).}
    \label{fig:quadtree_diagram}
    \vspace*{-0.75em}
\end{figure*}

By organizing objects or regions into a hierarchical quadtree structure, we can model an approximate priority based on spatial relevance. For instance, in aerial robotics we may wish to emphasize nearby terrain features or potential obstacles, or specific segmentation fields over the others for navigation. Similarly, in autonomous driving, closer agents such as pedestrians or vehicles demand more attention than those further away. Conveniently, quadtrees also split an image into a hierarchy of patches, which can be modeled as tokens in image transformers, making this representation well-suited for efficient perception in robotics.

To formalize this idea, we propose a hierarchical attention mechanism for autonomous perception called the {\em Sibling-Selective Quadtree Attention} (SSQA). Intuitively, we organize tokens into a quadtree hierarchy, where each parent cell aggregates information from its children. Within each sibling group, A lightweight {\em saliency-based Top-$k$ filtering} adaptively selects the most informative sibling interactions. This design provides a principled form of structured sparsity: attention cost grows {\em nearly linearly}, while interpretability is retained through explicit hierarchical grouping. Crucially, global receptive fields are preserved via communication between parent agents across levels, making the model both efficient and expressive.

In summary, the key contributions of this work include: 

\begin{itemize}
\item We propose \textbf{Sibling-Selective Quadtree Attention (SSQA)}, a hierarchical attention mechanism with constant-size sibling interactions and bottom--up aggregation (\S\ref{sec:method}).
\item We introduce \textbf{Cross-Injection Fusion}, a gated residual mechanism that injects parent evidence back to children, preserving suppressed information (\S\ref{sec:pairwise}).
\item We extend SSQA with \textbf{plug-in guidance}: (i) \textbf{Agent-Based SSQA (Image Field)} using detector/box-driven dynamic sparsity (\S\ref{sec:agent_image}), and (ii) \textbf{Agent-Based SSQA (Signal Field)} using Fourier-based refinement that focuses computation in high-frequency regions (\S\ref{sec:fourier}).
\item We integrate SSQA into a state-of-the-art end-to-end driving model (TransFuser\cite{chitta2022transfuser}) and achieve the highest average score on the CARLA benchmark across 14 scenarios against baseline and the other top down quadtree.
\end{itemize}

\section{Related}
\subsection{Efficient Attention in Vision Transformers}
The original Vision Transformer (ViT) showed that images can be tokenized into a fixed size patch grid and processed with global self-attention where each patch attends to one another, but the quadratic cost makes high-resolution tasks expensive~\cite{dosovitskiy2020vit}. %
DeiT~\cite{touvron2021deit} reduced data and compute via distillation and training strategies rather than architectural changes. Hierarchical backbones such as PVT~\cite{wang2021pvt} and Swin~\cite{liu2021swin} control cost via pyramids and shifted windows. For dense prediction, DPT couples a ViT backbone with multi-scale decoding to recover detail~\cite{ranftl2021dpt}. 

\subsection{Approximate, Sparse Attention}
Another line of work accelerates attention by either modifying the similarity kernel (the function that scores query–key pairs) or by imposing a sparsity pattern that limits which token pairs can interact, yielding near-linear scaling with the number of tokens. Linformer uses low-rank projections~\cite{wang2020linformer}; Performer kernelizes softmax via random features~\cite{choromanski2020performer}; Reformer uses ``locality sensitive hashing'' (LSH) to restrict matches~\cite{kitaev2020reformer}. Longformer and BigBird impose sliding-window and global tokens with theoretical guarantees~\cite{beltagy2020longformer,zaheer2020bigbird}; Sparse Transformers pioneered fixed strided patterns~\cite{child2019sparse}. FNet replaces attention with Fourier mixing~\cite{lee2021fnet}. These approximations trade some fidelity or flexibility for speed, which are effective for classification and dense tasks, but they may under-utilize fine spatial structures. In our work, we seek to maintain or even improve performance while optimizing modeling efficiency.

\subsection{Token Pruning and Merging}
A complementary strategy reduces the token count itself. DynamicViT prunes redundant tokens progressively and dynamically based on the input~\cite{rao2021dynamicvit}; EViT reorganizes tokens to emphasize salient regions~\cite{liang2022evit}; TokenLearner learns a small, content-adaptive set of tokens~\cite{ryoo2021tokenlearner}; ToMe merges similar tokens on-the-fly~\cite{bolya2023tome}; ATS performs content-aware sampling~\cite{yang2022ats}. These strategies reduce computations, yet still rely on global or windowed mixing among tokens.

\subsection{Quadtree and Hierarchical Data Structures}
Classical quadtrees recursively decompose space into quadrants based on informative regions~\cite{samet1984quadtree}. Quadtree Attention brings this idea into transformers by building token pyramids and restricting fine-level attention to regions selected at coarser levels, moving complexity from quadratic to linear, while retaining salient context~\cite{tang2022quadtree}. Our work follows this hierarchical principle, but differs in how selection and aggregation are organized, to be described in \S~\ref{sec:method}. In short, Quadtree Attention~\cite{tang2022quadtree} processes tokens in a top-down manner, where only top-$k$ salient patches are further subdivided into smaller resolutions. As a result, depending on the $k$ hyperparameter, it is possible that sparser details may be neglected with this top-down approach. Our model addresses this by proposing an alternative bottom-up gating approach with quadtrees with soft weighting, ensuring that all features have opportunity to be represented, even those with smaller pixel area. A visualization of this comparison can be found in Figure~\ref{fig:quadtree_diagram}.

\subsection{Structure-Guided Attention and Matching}
Task-driven evidence supports geometry-aware interaction constraints. SuperGlue attends over sparse keypoints with a graph-matching head~\cite{sarlin2020superglue}, while LoFTR achieves dense, detector-free correspondence by combining global context and local refinement~\cite{sun2021loftr}. These show that imposing structure on interactions can improve accuracy and efficiency, motivating our sibling-group attention and bottom-up fusion.

\subsection{Positioning and Differences from Prior Work}
Relative to windowed/backbone hierarchies~\cite{liu2021swin,wang2021pvt} and approximate/sparse kernels~\cite{wang2020linformer,choromanski2020performer,kitaev2020reformer,beltagy2020longformer,zaheer2020bigbird,child2019sparse,lee2021fnet}, our approach organizes attention \emph{within quadtree sibling groups}, where a lightweight saliency score weights the aggregation of children tokens in constant-size groups. Through aggregation of children token features, we form a compact \emph{parent} token representation, which is then \textit{fused} back to all siblings (bidirectional information exchange), before passing parent features upward to the next parent level. Unlike Quadtree Attention~\cite{tang2022quadtree}, which ranks across larger regions and goes deeper down into the trees, our selection is strictly local (per-sibling group) from a pre-constructed, and normally smaller tree depth compared to theirs, but with constant interaction size; our bottom-up bank and temperature-gated global fusion aggregate information across levels. With fixed sibling size, depth, and $k$, the number of interactions per level is constant, so the overall cost scales \emph{linearly} with the number of patches. 

\section{Methodology}
\label{sec:method}

We present \textbf{\emph{Sibling-Selective Quadtree Attention (SSQA)}} together with a \textbf{\emph{hierarchical agent-based self-attention}} module, where “agents” correspond to \textbf{objects} in the scene. SSQA works on its own, and can also consume agentness signals in two complementary forms: (i) \textbf{Agent-Based SSQA (Image Field)} using detector cues such as bounding boxes or SAM-derived heatmaps, and (ii) \textbf{Agent-Based SSQA (Signal Field)} using Fourier saliency maps obtained from FFT-based high-pass filtering. These signals bias local sibling interactions and drive optional adaptive refinement. At each level for SSQA, the patch grid is partitioned into $2{\times}2$ sibling groups; each child interacts only within their group via constant-size attention, updated children are summarized into a parent, and parents propagate information upward. This replaces global $N{\times}N$ interactions with within-quad attention and bottom--up aggregation. Throughout our main experiments, we use a fixed $2{\times}2$ pyramid with a \emph{soft top-$k$} mask over sibling logits (no external cues). The image-field and signal-field agentness variants are presented as optional extensions.%

\subsection{Notation}
\noindent\textbf{Patches and embedding.}
An input image is split into $N{=}\!h{\times}w$ patches (height $h$, width $w$). A ViT-style patchifier $\phi$ maps patches to
$\mathbf{Z}\!\in\!\mathbb{R}^{N\times D}$, where $D$ is the channel (feature) dimension.

\noindent\textbf{Heads.}
We use $H$ attention heads; per-head width is $d_h{=}D/H$.

\noindent\textbf{Hierarchy.}
A quadtree over an $h{\times}w$ grid halves spatial resolution per level $\ell=1,\dots,L$. 
Each level contains $G_\ell$ sibling groups of 4 children ($N_\ell=4G_\ell$). 
Totals are $S_G=\sum_\ell G_\ell$ and $S_N=4S_G$.

For the fixed $2{\times}2$ pyramid, 
$G_\ell=\frac{hw}{4^\ell}$ and $L=\lfloor\log_2\min(h,w)\rfloor$.

\noindent\textbf{Soft top-$k$.}
Within a $2{\times}2$ group, we keep the $k$ largest logits in each query row and \emph{clamp} the others to a small, data-dependent floor (e.g., mean minus a multiple of the row standard deviation) before softmax. We use $k\!\in\!\{2,3\}$ unless stated otherwise. During training, $\tau_{\mathrm{grp}}$ and $\tau_{\mathrm{pool}}$ are linearly annealed from 2.0 to 1.10.

\subsection{Patch Projection and Sibling Buffers for SSQA}
\label{sec:buffers}
Since we have $N{=}\!h{\times}w$ patches, $D$-dimensional features.
We compute $\mathbf{Z}=\phi(\text{image})$ and precompute \emph{sibling index buffers}. At level $\ell$, for each group $g$, the buffer lists the four child indices $\{i_{g,1},\dots,i_{g,4}\}$ so we can gather $\mathbf{X}\!\in\!\mathbb{R}^{4\times D}$ by pure indexing (no data-dependent control flow). We consider (i) a fixed pyramid using deterministic $2{\times}2$ tiling by repeated integer halving, yielding
$L=\lfloor\log_2\min(h,w)\rfloor$ and $G_\ell=\tfrac{hw}{4^\ell}$,
and (ii) an adaptive variant where detector-/Fourier-guided saliency alters per-image depth and group counts $(G'_\ell)$; buffers can be cached.

\subsection{Pairwise Sibling Attention with Cross-Injection Fusion for SSQA}
\label{sec:pairwise}

At hierarchy level $\ell$, consider one sibling group $g$ with its four child tokens stacked as
$\mathbf{X}\in\mathbb{R}^{4\times D}$ (rows are children; $D$ is the feature dimension).
We use $H$ attention heads, each of width $d_h{=}D/H$.

\paragraph{One shared projection for Q, K, V.}
We apply a single linear projection 
$\mathbf{W}_{qkv}\in\mathbb{R}^{D\times 3D}$ 
to obtain
\[
\mathbf{Q},\mathbf{K},\mathbf{V}\in\mathbb{R}^{4\times D}.
\]

\paragraph{Within-quad attention (4$\times$4, per head).}
Split $\mathbf{Q},\mathbf{K},\mathbf{V}\in\mathbb{R}^{4\times D}$ into $H$ heads:
$\mathbf{Q}^{(h)},\mathbf{K}^{(h)},\mathbf{V}^{(h)}\in\mathbb{R}^{4\times d_h}$ with $d_h=D/H$ and $h=1,\dots,H$.
For each head, the scaled dot-product logits are
\[
\mathbf{A}^{(h)}[i,j] \;=\; \frac{\langle \mathbf{Q}^{(h)}_{i,:},\,\mathbf{K}^{(h)}_{j,:}\rangle}{\sqrt{d_h}},
\quad i,j\in\{1,2,3,4\}.
\]

\noindent\textit{Soft top-$k$ (default).}
For each head, we retain the $k$ largest logits in every query row of $\mathbf{A}^{(h)}$ and clamp the others to a small data-dependent floor. The masked logits are scaled by $\tau_{\mathrm{grp}}$ and stabilized before row-wise softmax normalization. Top-$k$ entries dominate while gradients remain nonzero for all siblings.

\noindent Next, mix values within the group (still per head),
\[
\mathbf{M}^{(h)} \;=\; \mathbf{W}^{(h)}\,\mathbf{V}^{(h)} \in \mathbb{R}^{4\times d_h},
\]
\noindent so each child’s updated representation is a weighted blend of the four value vectors for that head (a convex combination per row).

\noindent Finally, concatenate heads along the channel dimension to get the updated children:
\[
\widehat{\mathbf{X}} \;=\; \operatorname{Concat}_{h=1}^H \mathbf{M}^{(h)} \in \mathbb{R}^{4\times D}.
\]
\noindent Concatenation restores the full $D$ channels, yielding four updated child tokens. All communication was \emph{local} within a $2{\times}2$ group—no global $N{\times}N$ mixing—and the per-head attention remained $4{\times}4$, giving constant compute per group.

\paragraph{Cross-Injection Fusion}
At this stage, we have the original children $\mathbf{X}\in\mathbb{R}^{4\times D}$
and their attention-updated versions $\widehat{\mathbf{X}}\in\mathbb{R}^{4\times D}$.
To avoid blindly replacing one with the other, we blend them with a
\emph{gated residual}:
\[
\widetilde{\mathbf{X}}=\mathbf{X}+\gamma\,\sigma(\mathbf{X}\mathbf{W}_{g})\odot(\widehat{\mathbf{X}}-\mathbf{X}),
\]
where $\mathbf{W}_g\in\mathbb{R}^{D\times D}$ is a shared projection, $\sigma(\cdot)$ is a sigmoid, and $\gamma$ is a learned scalar constrained to a bounded interval for numerical stability. Each child learns per-channel gates telling how much of the update to accept
(open the gate) or to ignore (close the gate). Let $\widetilde{\mathbf{x}}_c \in \mathbb{R}^D$ denote the $c$-th row of $\widetilde{\mathbf{X}}$.

\paragraph{Lightweight per-child MLP.}
Each child is then refined independently with the same two-layer feed-forward network:
LayerNorm $\to$ Linear($D\!\to\!3D$) $\to$ GELU $\to$ Dropout $\to$ Linear($3D\!\to\!D$) $\to$ Dropout.
The output is added back via a scaled residual connection,
\[
\widetilde{\mathbf{X}} \leftarrow 
\widetilde{\mathbf{X}} + \eta\,\mathrm{MLP}(\mathrm{LN}(\widetilde{\mathbf{X}})),
\]
where $\eta=0.5$ is a fixed residual scaling coefficient.
This acts like a small local transformer block at the sibling level, adding nonlinearity and channel mixing without high cost.

\paragraph{Parent and summary.}
Finally, we summarize the group by averaging its four children:
\[
\mathbf{h}^{(\ell)}_{g}=\tfrac{1}{4}\sum_{c=1}^{4}\mathrm{LN}(\widetilde{\mathbf{x}}_{c}).
\]
This summary is cached for global gating (Sec.~\ref{sec:gating}),
and—after LayerNorm and a linear projection—becomes the \emph{parent token}
that moves up to the next level:
\[
\mathbf{o}_p=\mathrm{Linear}(\mathrm{LN}(\mathbf{h}^{(\ell)}_{g})).
\]
The resulting parent token serves as the input representation at level $\ell+1$.

\subsection{Bottom--Up Aggregation and Global Gating for SSQA}
\label{sec:gating}
We use a pooling temperature $\tau_{\mathrm{pool}}\ge 1.10$ for global aggregation.
Let $T=\sum_{\ell=1}^{L} G_\ell$ denote the total number of sibling groups across all levels.
We concatenate all group summaries $\mathbf{h}^{(\ell)}_g$ into
\[
\mathbf{F}\in\mathbb{R}^{T\times D},
\]
where each row is the $D$-dimensional representation of one sibling group.

We then compute a set of scalar importance scores:
\begin{align*}
\boldsymbol{\pi}&=\mathbf{F}\mathbf{w}_p \in \mathbb{R}^{T},\qquad
\mathbf{w}=\operatorname{softmax}\!\bigl(\boldsymbol{\pi}/\tau_{\mathrm{pool}}\bigr),
\end{align*}
where $\mathbf{w}_p\in\mathbb{R}^{D\times 1}$ is a learned projection that compresses each group vector into a single score. The vector $\boldsymbol{\pi}$ contains these raw scores, and the softmax turns them into a probability distribution $\mathbf{w}$ that highlights the most influential groups while still keeping every group nonzero.

Next we form a global summary vector:
\begin{align*}
\mathbf{r}&=\sum_{t=1}^{T} w_t\,\mathbf{F}_t \;+\; \lambda\,\tfrac{1}{T}\sum_{t=1}^{T}\mathbf{F}_t,
\end{align*}
which is a weighted average of group features according to importance weights $w_t$, plus a small uniform residual (with $\lambda=0.1$) to prevent the model from ignoring weaker groups completely. Intuitively, $\mathbf{r}$ is the single “consensus” feature vector that carries information from all quads, with more focus on the ones judged important.

Finally, a linear classifier maps $\mathbf{r}$ to task logits.

\begin{algorithm}[H]
\caption{SSQA (one forward pass; optional guided/FFT hook)}
\label{alg:ssqa}
\begin{algorithmic}[1]
\STATE $\mathbf{Z}\leftarrow\phi(\text{image})$ \COMMENT{ViT patch embedding}
\STATE Precompute quad groups $\{\mathcal{G}^{(\ell)}\}_{\ell=1}^{L}$
\FOR{$\ell=1$ \TO $L$}
  \FOR{each group $g\in\mathcal{G}^{(\ell)}$}
    \STATE $\mathbf{X}\in\mathbb{R}^{4\times D}$ \COMMENT{gather siblings}
    \STATE $\bar{\mathbf{X}}\leftarrow \mathrm{LN}(\mathbf{X})$ \COMMENT{pre-LN}
    \STATE $[\mathbf{Q},\mathbf{K},\mathbf{V}]\leftarrow \bar{\mathbf{X}}\mathbf{W}_{qkv}$ \COMMENT{split into $H$ heads}
    \STATE $\mathbf{A}\leftarrow \mathbf{Q}\mathbf{K}^{\top}/\sqrt{d_h}$ \hfill(+ optional bias)
    \STATE \textbf{soft top-$k$}: row-wise keep top-$k$, clamp others, scale by $\tau_{\mathrm{grp}}$, subtract row max
    \STATE $\widehat{\mathbf{X}}\leftarrow \mathrm{softmax}(\mathbf{A})\,\mathbf{V}$ \COMMENT{within-quad attention}
    \STATE $\mathbf{G}\leftarrow\sigma(\mathbf{X}\mathbf{W}_g)$
    \STATE $\widetilde{\mathbf{X}}\leftarrow \mathbf{X}+\gamma\,\mathbf{G}\odot(\widehat{\mathbf{X}}-\mathbf{X})$ \COMMENT{CIF}
    \STATE $\widetilde{\mathbf{X}}\leftarrow \widetilde{\mathbf{X}}+\eta\,\mathrm{MLP}(\mathrm{LN}(\widetilde{\mathbf{X}}))$
    \STATE $\mathbf{h}^{(\ell)}_g\leftarrow \tfrac{1}{4}\sum_{c}\mathrm{LN}(\widetilde{\mathbf{x}}_c)$; $\mathbf{o}_p\leftarrow \mathrm{Linear}(\mathbf{h}^{(\ell)}_g)$
  \ENDFOR
  \STATE children $\leftarrow$ parents $\{\mathbf{o}_p\}$
\ENDFOR
\STATE $\mathbf{F}\leftarrow \mathrm{concat}(\{\mathbf{h}^{(\ell)}_g\})$
\STATE $\mathbf{w}\leftarrow \mathrm{softmax}((\mathbf{F}\mathbf{w}_p)/\tau_{\mathrm{pool}})$
\STATE \textbf{return} $\mathrm{head}\!\left(\sum_t w_t\mathbf{F}_t + \lambda\,\tfrac{1}{T}\sum_t \mathbf{F}_t\right)$
\end{algorithmic}
\end{algorithm}

\subsection{Agent-Based SSQA (Image Field)}
\label{sec:agent_image}
We first try to bias SSQA toward \emph{agents} via a single heatmap $\mathbf{H}\!\in\![0,1]^{h\times w}$ aligned with the patch grid (each cell corresponds to one image patch). 
The heatmap can be constructed from \textbf{masks/saliency} (e.g., SAM masks, merged and downsampled to the patch grid) or \textbf{detector boxes} (e.g., YOLO bounding boxes rasterized to grid cells; optional with higher overhead). Both constructions operate in the spatial (image) domain.

\paragraph{Agent-biased sparsity.}
For each $2{\times}2$ sibling group $g$, let $h_{g,1..4}$ denote the corresponding heatmap values. 
We compute a group score
\[
s_g=\alpha\,\max(h_{g,1..4})+(1-\alpha)\,\mathrm{mean}(h_{g,1..4}),
\]
and determine the retained sibling count
\[
k_g = k_{\min} + \left\lfloor (k_{\max}-k_{\min})\, s_g^\gamma \right\rfloor,
\qquad 1 \le k_g \le 4.
\]
Attention within group $g$ is restricted to the top-$k_g$ children.

\paragraph{Adaptive refinement.}
Optionally, we adapt the quadtree by subdividing nodes whose heatmap mass exceeds a threshold $\tau_H$ up to depth $L_{\max}$, yielding spatially adaptive refinement.

\subsection{Agent-Based SSQA (Signal Field)}
\label{sec:fourier}

We derive a Fourier high-pass saliency map and pool it to the patch grid,
yielding spectral scores $\mathbf{H}$ (flattened as $\mathbf{h}$).

\paragraph{HP-biased Selection and Pooling.}
We bias within-group logits and global gating as
\begin{align}
s_c &\leftarrow s_c^{\text{attn}} + \alpha_{\text{hp}}\, h_c, \\
\boldsymbol{\pi} &\leftarrow \mathbf{F}\mathbf{w}_p + \beta\, \bar{\mathbf{h}},
\end{align}
where $h_c$ denotes the spectral response of child $c$ and $\bar{\mathbf{h}}$ aggregates group-wise high-pass scores.

\paragraph{HP-guided Adaptive Buffers.}
We optionally construct adaptive pyramids by subdividing cells with large high-pass statistics until depth $L_{\max}$, reducing active groups in smooth regions.

\subsection{Complexity Analysis for SSQA}
\label{sec:complexity}

\paragraph{Preprocessing}
Patch embedding costs $O(ND)$. Fixed buffers are built once by integer halving ($O(L\,hw)$).
The image-field agent map (Sec.~\ref{sec:agent_image}) is obtained once per image from external detectors or mask generators (e.g., YOLO boxes or SAM masks) and cached on the patch grid. 
The Fourier map (Sec.~\ref{sec:fourier}) costs $O(224^2\log 224)$ for 2-dimensional FFT and $O(224^2)$ for pooling; it is also performed once per image and cached.

\paragraph{Forward}
At level $\ell$:
\begin{itemize}
\item QKV projections on $N_\ell$ children: $3D^2 N_\ell$ MACs.
\item CiF gate projection per child: $D^2 N_\ell$ MACs.
\item Local two-layer MLP per child: $6D^2 N_\ell$ MACs.
\item Sibling attention (constant-size $4{\times}4$ per group): $32D\,G_\ell$ MACs (QK and KV each $16D\,G_\ell$).
\item Parent projection: $D^2 G_\ell$ MACs.
\end{itemize}
Summing over levels,
\begin{align*}
\text{MACs}_{\text{SSQA}}
&= \sum_{\ell=1}^L 
   \bigl(10D^2 N_\ell + D^2 G_\ell + 32D\,G_\ell\bigr)\\
   + O(D\,S_G).
\end{align*}
When $h,w$ are powers of two, $N_\ell{=}N/4^{\ell-1}$ and $G_\ell{=}N/4^\ell$ so that $S_N{=}\tfrac{4}{3}N$ and $S_G{=}\tfrac{1}{3}N$. Plugging in yields
\begin{align*}
\text{MACs}_{\text{SSQA}}
= \Bigl(\tfrac{40}{3}\,D^2 N\Bigr)
+ \Bigl(\tfrac{32}{3}\,D N\Bigr)
+ O(DN)
\end{align*}

The dominant term is $O(D^2N)$; the quadratic $O(N^2D)$ term of global attention is eliminated.

\section{Results}

\subsection{Overview.}
\begin{figure}[h]
\centering
\begin{subfigure}{0.8\linewidth}
  \includegraphics[width=\linewidth]{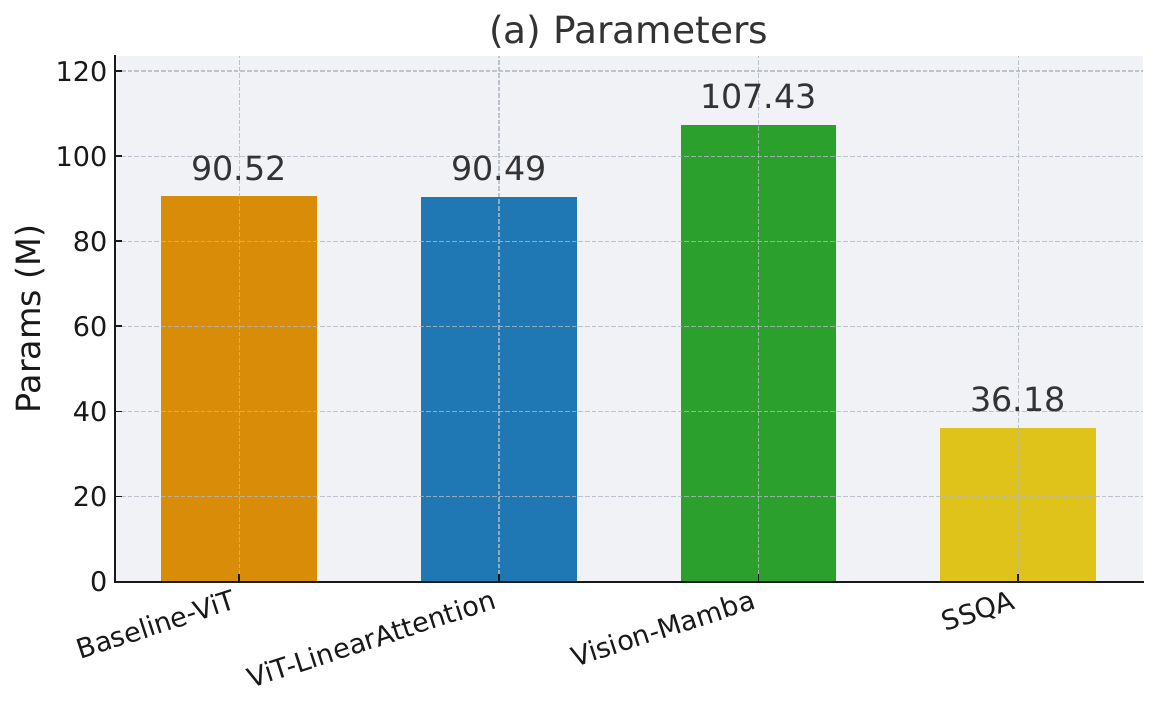}
  \caption{Parameters}
\end{subfigure}\hfill
\begin{subfigure}{0.8\linewidth}
  \includegraphics[width=\linewidth]{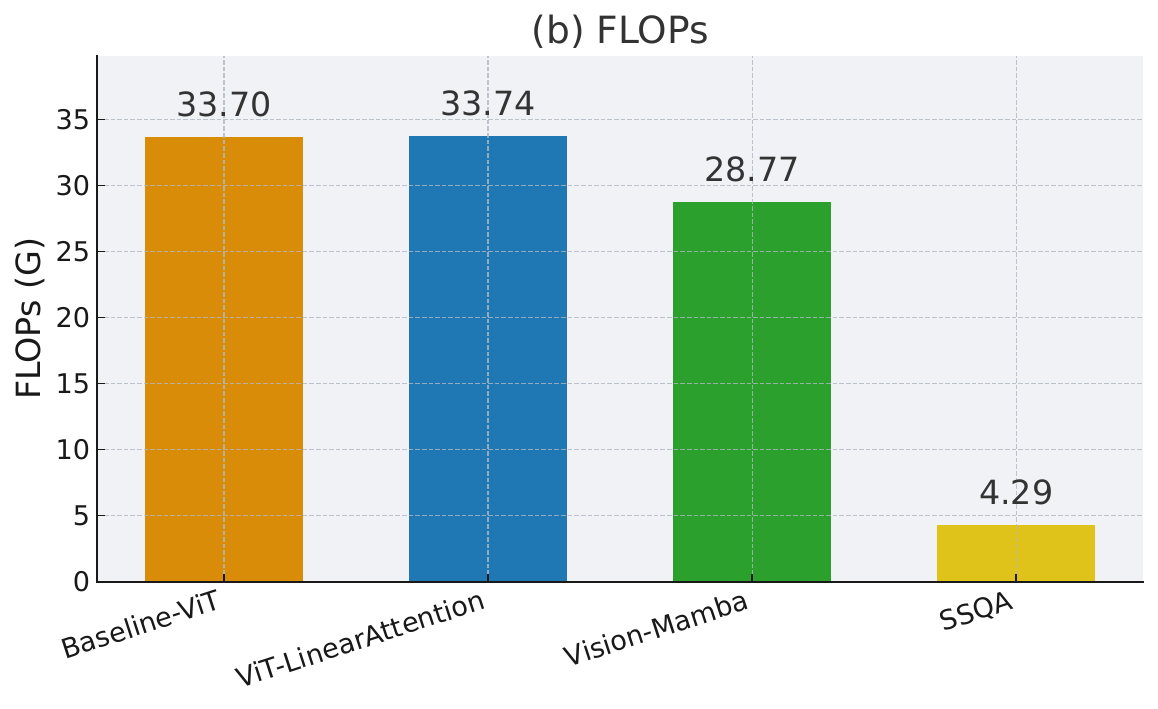}
  \caption{FLOPs}
\end{subfigure}
\caption{{\bf Model Efficiency}. SSQA is substantially lighter by reducing the number of parameters by nearly {\bf 3x} and more compute-efficient by up to a factor of {\bf 8x} for classification test.}
\label{fig:efficiency}
\end{figure}

We evaluate \textbf{SSQA} on image classification, semantic segmentation, and end-to-end driving in CARLA.
Across tasks, SSQA matches or improves accuracy while using substantially fewer parameters and FLOPs.
We report training dynamics (Fig.~\ref{fig:training_dynamics}), efficiency (Fig.~\ref{fig:efficiency}),
CARLA~\cite{dosovitskiy2017carla} performance (Figs.~\ref{fig:carla_score}), and per-task summaries (Tabs.~\ref{tab:cls}--\ref{tab:seg}).

\subsection{Experimental Setup}
\textbf{Benchmarks.} We use three datasets: \emph{IllusionAnimals} for binary classification under heavy camouflage and clutter,
\emph{LoveDA} for land-cover semantic segmentation relevant to robotic navigation (e.g., region understanding for safe landing),
and \emph{CARLA}~\cite{dosovitskiy2017carla} for end-to-end driving over 14 benchmark scenarios (higher score is better).

\textbf{Training.} All runs use a single NVIDIA A100 GPU and AdamW ($3\times10^{-4}$, wd $0.05$, $\beta=(0.9,0.95)$).
We train IllusionAnimals for 50 epochs (bs=32, $224^2$) with a ViT-16 baseline and prior efficient backbones;
LoveDA is trained for 50 epochs (bs=12, $512^2$) using a ViT-16 style baseline (patch=16, depth=6, heads=8) and the same setting for all models.
We use BCEWithLogits for classification and cross-entropy for segmentation. Images are normalized with ImageNet statistics. Unless otherwise noted, we report single-seed results; for key ablations we run three seeds (42/43/44) and report mean$\pm$std.

\textbf{Guidance and compute.} Guided variants use YOLO boxes (classification) or SAM masks (segmentation).
FLOPs are measured with THOP using $1\times3\times224\times224$ (cls) and $1\times3\times512\times512$ (seg) inputs.

\subsection{Binary Image Classification (IllusionAnimals)}

\begin{table}[h]
\caption{{\bf Compute vs.\ Validation performance} for image classification: upto {\bf 3x} fewer parameters, {\bf 8x} fewer FLOPs, {\bf 2.08x} mean epoch time.}
\label{tab:cls}
\centering
\scriptsize
\setlength{\tabcolsep}{4pt}
\renewcommand{\arraystretch}{1.1}
\begin{tabular}{lcccc}
\toprule
\textbf{Method} & \textbf{Params (M)$\downarrow$} & \textbf{FLOPs (G)$\downarrow$} & \textbf{Acc$\uparrow$} & \textbf{Avg Time (s) $\downarrow$}\\
\midrule
Baseline-ViT~\cite{dosovitskiy2020image}  & 90.52  & 33.70 & 98.25\% & 53.17\\
ViT-LinAtt~\cite{katharopoulos2020transformers}  & 90.49  & 33.74 & 98.25\% & 54.65\\
Mamba 2~\cite{gu2023mamba}         & 107.43 & 28.77 & 98.41\% & 67.70\\
\textbf{SSQA (ours)} & \textbf{36.18} & \textbf{4.29} & \textbf{98.65\%} & \textbf{26.19}\\
\bottomrule
\end{tabular}
\end{table}

\begin{figure}[h]
\centering
\begin{subfigure}{0.9\linewidth}
  \includegraphics[width=\linewidth]{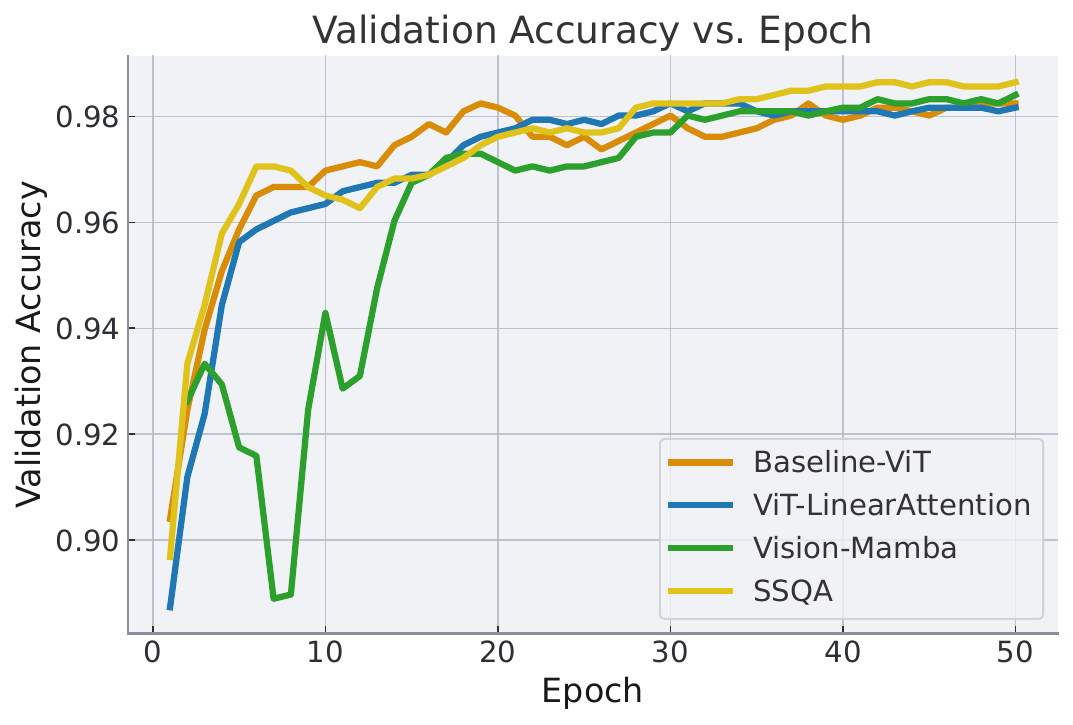}
  \caption{Val Acc vs.\ Epochs}
\end{subfigure}\hfill
\begin{subfigure}{0.9\linewidth}
  \includegraphics[width=\linewidth]{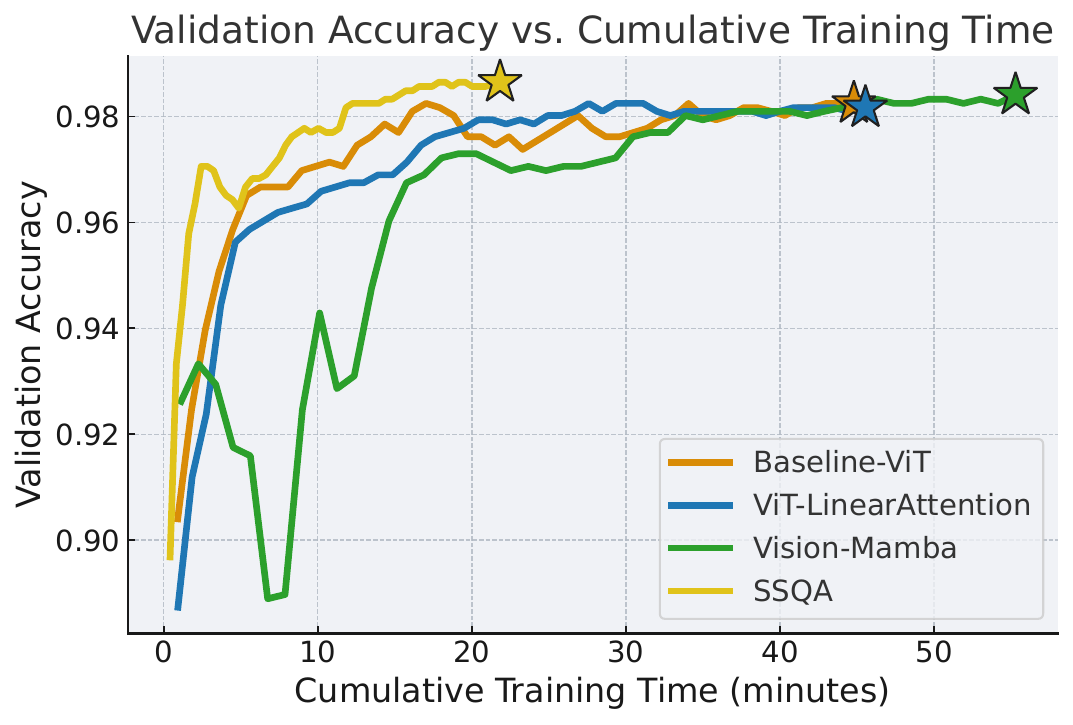}
  \caption{Acc vs.\ Cum.\ Time}
\end{subfigure}
\caption{{\bf Training Dynamics}. SSQA (a) converges faster and (b) reaches target accuracy earlier than the ViT baseline.}
\label{fig:training_dynamics}
\end{figure}

\begin{table}[h]
\caption{{\bf Thorough Compute vs.\ Accuracy} for Classification on IllusionAnimals (50 epochs, $224^2$).}
\label{tab:cls50}
\centering
\footnotesize
\setlength{\tabcolsep}{4pt}
\renewcommand{\arraystretch}{1.1}
\begin{tabular}{lccc}
\toprule
\textbf{Method} & \textbf{Params (M)} & \textbf{FLOPs (G)$\downarrow$} & \textbf{Acc} \\
\midrule
Baseline-ViT~\cite{dosovitskiy2020image}                & 90.52 & 33.70 & 98.25\% \\
ViT-LinAtt~\cite{katharopoulos2020transformers}         & 90.51 & 33.74 & 98.25\% \\
Mamba 2~\cite{dao2024transformersssm}                   & 107.43 & 28.77 & 98.41\% \\
Swin TRXFM~\cite{liu2021swin}                           & 32.24 & 8.75 & 99.05\% \\
MLP-Mixer~\cite{tolstikhin2021mlpmixer}                 & 63.84 & 25.24 & 98.02\% \\
ConvNeXT V2~\cite{woo2023convnextv2codesigningscaling}  & 32.54 & 8.92 & 98.33\% \\
Quadtree ATTN-B2~\cite{tang2022quadtree}                & \textbf{27.63} & 4.50 & 98.49\% \\
Quadtree ATTN-B3~\cite{tang2022quadtree}                & 46.30 & 7.29 & \textbf{99.50\%} \\
\midrule
\textbf{SSQA (ours)}                                    & 36.18 & 4.29 & 98.65\% \\
SSQA-Guided-YOLO (ours)                                 & 36.47 & 4.41 & 98.41\% \\
SSQA-FFT (ours)                                         & 37.20 & \textbf{3.71} & 98.20\% \\
\bottomrule
\end{tabular}
\end{table}

\noindent\textbf{Comparison.}
\textbf{SSQA} achieves the best validation accuracy while converging faster than the ViT baseline (Fig.~\ref{fig:training_dynamics}).
With \textbf{36.18\,M} parameters and \textbf{4.29\,G} FLOPs, it reaches \textbf{98.65\%} accuracy,
reducing parameters by up to \textbf{3$\times$}, FLOPs by up to \textbf{8$\times$}, and mean epoch time by \textbf{2.08$\times$} (Tab.~\ref{tab:cls}).

Compared to ViT~\cite{dosovitskiy2020image} and Mamba-2~\cite{gu2023mamba,dao2024transformersssm},
SSQA uses $\sim$\textbf{8$\times$} and $\sim$\textbf{6.7$\times$} fewer FLOPs, respectively, while achieving higher accuracy.
It remains within \textbf{0.40} percentage points of Swin~\cite{liu2021swin} at roughly \textbf{2$\times$} lower compute.
Relative to Quadtree Attention~\cite{tang2022quadtree}, SSQA outperforms B2 (98.49\%) at similar FLOPs and is more efficient than B3,
which requires \textbf{1.7$\times$} more FLOPs for higher peak accuracy.
Overall, SSQA provides the strongest accuracy--compute trade-off (Tab.~\ref{tab:cls50}).

\noindent\textbf{Stability (multi-seed).}
With the default $2\times2$ sibling grouping, SSQA achieves $98.65\pm0.03\%$ accuracy and $0.9974\pm0.0012$ AUC over 3 seeds (42/43/44), indicating low variance under the same training recipe.

\subsection{Ablation Study (IllusionAnimals)}
\label{subsec:ablation}

\begin{table}[h]
\caption{{\bf Ablation on SSQA components} (IllusionAnimals, 50 epochs, $224^2$). We report best validation accuracy for one run (seed=42) and include compute for reference; key ablations additionally report mean$\pm$std over 3 seeds (42/43/44) when available.}
\label{tab:ablation}
\centering
\footnotesize
\setlength{\tabcolsep}{4pt}
\renewcommand{\arraystretch}{1.1}
\begin{tabular}{lcccc}
\toprule
\textbf{Variant} & \textbf{Params (M)} & \textbf{FLOPs (G)} & \textbf{Acc} & \textbf{$\Delta$Acc} \\
\midrule
SSQA (full; $k{=}3$) & 36.18 & 4.29 & 98.65\% & +0.00 \\
w/o CIF (OFF) & 36.18 & 4.29 & 98.22\% & $-0.43$ \\
NO-GATE (no CIF gate) & 36.18 & 4.29 & 98.53\% & $-0.12$ \\
w/o local FFN & 36.18 & 1.52 & 98.29\% & $-0.36$ \\
$k{=}2$ & 36.18 & 4.29 & 98.40\% & $-0.25$ \\
$k{=}4$ (no top-$k$) & 36.18 & 4.29 & 98.37\% & $-0.28$ \\
$\lambda{=}0$ & 36.18 & 4.29 & 98.17\% & $-0.48$ \\
\bottomrule
\end{tabular}
\end{table}

\textbf{Findings.}
Removing the lightweight local FFN reduces peak validation accuracy by $0.36$ pp while also lowering FLOPs, suggesting that local nonlinearity provides a modest but consistent gain beyond sibling attention.
Varying sparsity indicates that moderate sparsification ($k{=}3$) performs best: stronger sparsity ($k{=}2$) and fully dense mixing ($k{=}4$) reduce accuracy by $0.25$ pp and $0.28$ pp. This is expected as SSQA emphasizes on focusing on relatively important grids, while not all of them. Finally, disabling the uniform residual in global pooling ($\lambda{=}0$) reduces accuracy by $0.48$ pp, supporting its role in stabilizing aggregation across pyramid levels.

\subsection{Semantic Segmentation (LoveDA)}
\FloatBarrier

\begin{table}[h!]
\caption{{\bf Compute vs.\ mIoU} for Segmentation on LoveDA ($512^2$)}
\label{tab:seg}
\centering
\footnotesize
\setlength{\tabcolsep}{4pt}
\renewcommand{\arraystretch}{1.1}
\begin{tabular}{lccc}
\toprule
\textbf{Method} & \textbf{Params (M)} & \textbf{FLOPs (G)$\downarrow$} & \textbf{mIoU} \\
\midrule
Baseline-ViT~\cite{dosovitskiy2020image}& 24.94 & 49.63 & 0.3349 \\
ViT-LinAtt~\cite{katharopoulos2020transformers}                  & 23.98 & 43.22 & 0.2932 \\
Mamba 2~\cite{dao2024transformersssm}                     & 14.11 & 48.22 & 0.3518 \\
Swin TRXFM~\cite{liu2021swin}                  & 35.49 & 53.60 & 0.3307 \\
MLP-Mixer~\cite{tolstikhin2021mlpmixer}                   & 74.72 & 148.15& 0.3432 \\
ConvNeXT V2~\cite{woo2023convnextv2codesigningscaling}          & 33.13 & 24.64 & 0.2803 \\
Quadtree ATTN-B0~\cite{tang2022quadtree}               & \textbf{11.08} & 21.48 & \textbf{0.3563} \\ 
\midrule
\textbf{SSQA (ours)}        & 11.28 & \textbf{13.96} & 0.3547 \\
SSQA-Guided-SAM (ours)          & 12.28 & 14.25 & 0.3398 \\
SSQA-FFT (ours)             & 12.61 & 15.07 & 0.3151 \\

\bottomrule
\end{tabular}
\end{table}

\textbf{Comparison}
\textbf{SSQA} has the significantly lowest flop, and it also offers the strongest accuracy--compute trade-offs among segmentation models. 
With 13.96\,G FLOPs and 11.28\,M params, it attains \textbf{0.3547} mIoU while using \textbf{$\sim$3.5$\times$} fewer FLOPs 
and \textbf{$\sim$2$\times$} fewer parameters than the ViT baseline~\cite{dosovitskiy2020image}. 
Compared to Mamba-2~\cite{dao2024transformersssm} (0.3518 mIoU at 48.22\,G FLOPs), SSQA achieves higher accuracy with 
\textbf{$\sim$3.5$\times$} lower compute. 
Relative to Swin~\cite{liu2021swin} (0.3307 mIoU at 53.60\,G) and MLP-Mixer~\cite{tolstikhin2021mlpmixer} 
(0.3432 mIoU at 148.15\,G), SSQA surpasses both while requiring dramatically less compute (4--10$\times$ fewer FLOPs). 
Although Quadtree Attention~\cite{tang2022quadtree}'s mIOU is 0.16 percentage better, but it takes
\textbf{1.5$\times$} flops. Also, B0 is already the lightest tree they have provided, but we still have significantly fewer flops in this task for SSQA.
Overall, SSQA delivers near-top segmentation performance at substantially lower computational cost (Table~\ref{tab:seg}). 

\begin{figure}[h!]
    \centering
    \includegraphics[width=1\linewidth]{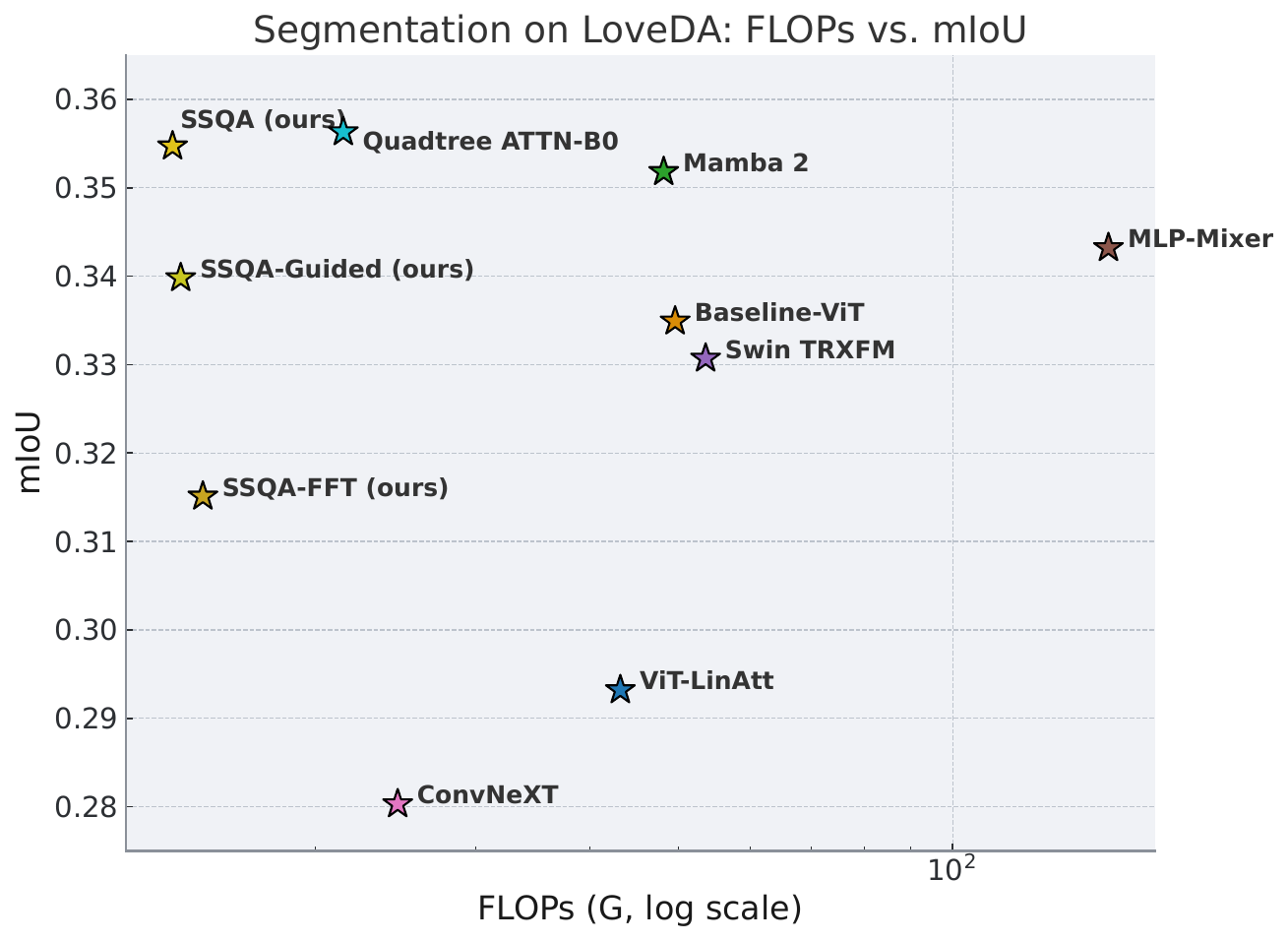}
    \caption{{\bf Segmentation Model Performance Comparison}. SSQA has the lowest FLOPs {\bf 1.54x} lower than the second lowest flops model and still be able to achieve the second best accuracy with only 0.16 percentage apart. }
    \label{fig:placeholder}
\end{figure}

\subsection{End-to-End Driving (CARLA Simulator)}
\label{subsec:carla}

\begin{figure}[t]
  \centering
  \includegraphics[width=1\linewidth]{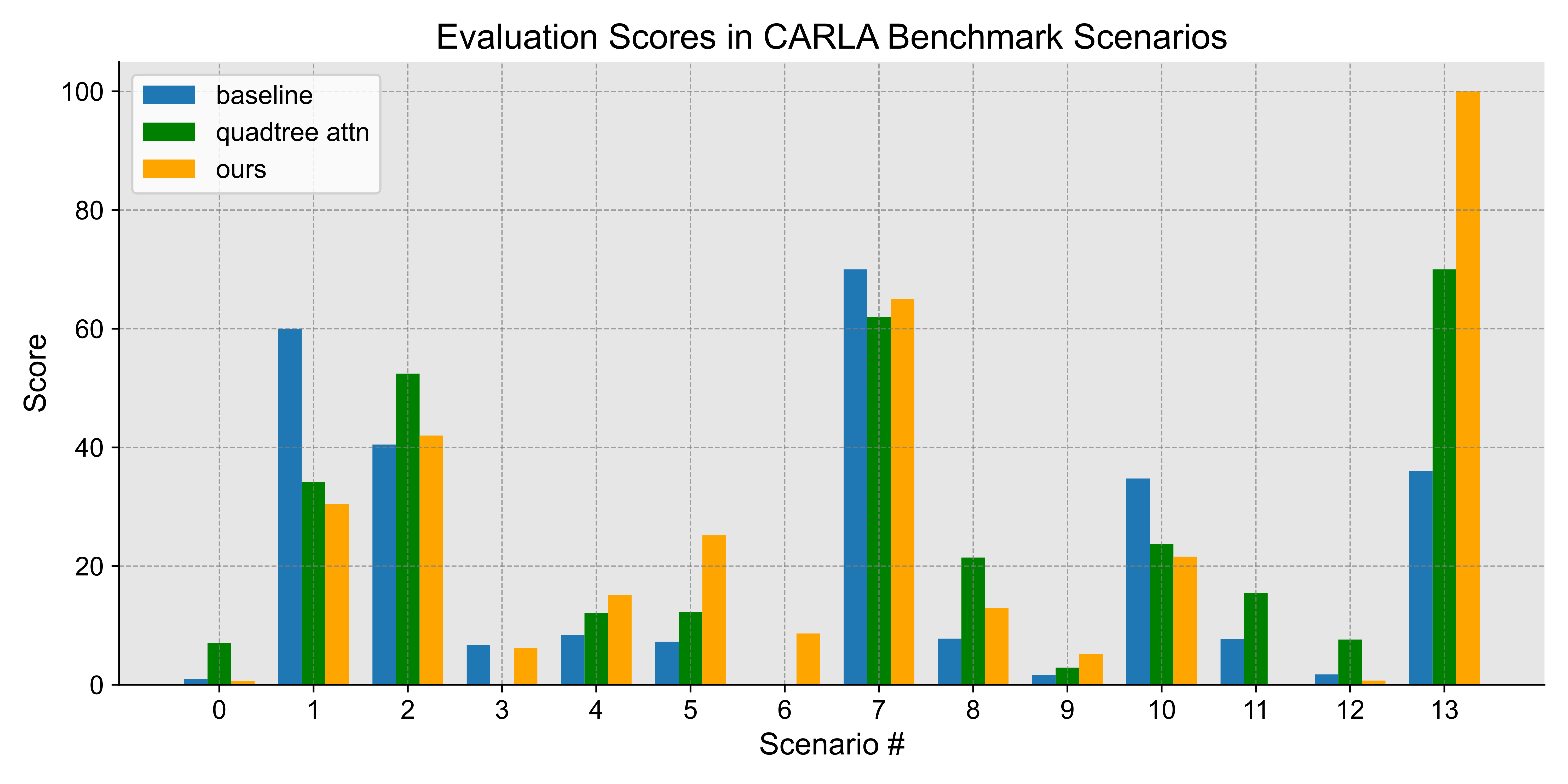} 
  \caption{{\bf CARLA Evaluation Score}. Benchmark evaluation score across 14 scenarios (0--13) for Baseline ~\cite{dosovitskiy2017carla} vs. Quadtreen ATTN~\cite{tang2022quadtree} vs. \ SSQA (higher is better).}
  \label{fig:carla_score}
\end{figure}

\textbf{Results}
Figure~\ref{fig:carla_score} compares the evaluation scores (higher is better) across 14 CARLA scenarios. Overall, SSQA achieves the highest average score (23.83), outperforming Quadtree Attention (22.95)~\cite{tang2022quadtree} and the baseline (20.25)~\cite{dosovitskiy2017carla}. SSQA also attains the largest number of scenario-wise wins. Notably, SSQA substantially improves Scenario 13 (36 $\rightarrow$ 100) and achieves strong gains in Scenarios 4, 5, and 6. In several scenarios (e.g., 2 and 7), SSQA remains competitive with both competing methods. However, it underperforms the baseline in a few cases, such as Scenarios 1 and 10. These results indicate that SSQA delivers consistent improvements across challenging scenarios, leading to the best overall mean performance.

\FloatBarrier

\subsection{Agent-Based Extension Evaluation}
\label{subsec:agent_ext}

The YOLO-, SAM-, and Fourier-guided variants introduce minimal computational overhead relative to SSQA. On IllusionAnimals, YOLO-guided SSQA achieves $98.41\%$ accuracy compared to $98.65\%$ for the base model ($-0.24$\,pp). The Fourier gate reduces FLOPs by emphasizing coarse global structure, but slightly degrades accuracy. On LoveDA, external priors provide negligible benefit under our training protocol.

We attribute these results to three factors. 
(1) \textbf{Prior mismatch:} bounding-box or region-based priors may suppress informative context in datasets with diverse object scales and textures. 
(2) \textbf{Over-regularization:} when SSQA already localizes effectively, additional gating may limit beneficial long-range interactions. 
(3) \textbf{Frequency bias:} emphasizing low-frequency structure can reduce boundary precision, explaining the FLOP–accuracy trade-off observed with Fourier gating. Overall, SSQA performs strongly without external guidance. 

\section{Conclusion}
In this paper we introduced Sibling-Selective Quadtree Attention (SSQA), a simple hierarchical self-attention layer built on a full quadtree. SSQA groups patches into fixed $2\times2$ siblings, attends only within each group, summarizes children into a parent, and feeds parent evidence back to children (Cross-Injection Fusion). The layer uses standard transformer operations, without special kernels or specific optimization, and integrates easily into existing backbones.
This effective hierarchical attention mechanism reduces the computational cost on machine perception tasks and scales linearly with the number of patches, while still allowing information to move across the images. In experiments, SSQA matches or improves accuracy with a much lower compute cost across classification (IllusionAnimals) and segmentation (LoveDA). 

\section*{Acknowledgements}
This work is supported in part by NSF/UMD REU-CAAR Program, Dr. Barry Mersky and 
Capital One E-Nnovte Endowed Professorships, University of Maryland Distinguished University Professorship, Maryland Transportation Institute Graduate Research 
Fellowship, National Science Foundation, and ARL-UMD ArtIAMAS Cooperative Agreement.

\bibliographystyle{ieeetr}
\bibliography{references}

\begin{thebibliography}{10}

\bibitem{liu2021swin}
Z.~Liu, Y.~Lin, Y.~Cao, H.~Hu, Y.~Wei, Z.~Zhang, S.~Lin, and B.~Guo, ``Swin
  transformer: Hierarchical vision transformer using shifted windows,'' in {\em
  Proceedings of the IEEE/CVF International Conference on Computer Vision
  (ICCV)}, 2021.

\bibitem{xie2021segformer}
E.~Xie, W.~Wang, Z.~Yu, A.~Anandkumar, J.~M. Alvarez, and P.~Luo, ``Segformer:
  Simple and efficient design for semantic segmentation with transformers,'' in
  {\em Neural Information Processing Systems (NeurIPS)}, 2021.

\bibitem{katharopoulos_et_al_2020_linear_transformers}
A.~Katharopoulos, A.~Vyas, N.~Pappas, and F.~Fleuret, ``Transformers are rnns:
  Fast autoregressive transformers with linear attention,'' in {\em Proceedings
  of the International Conference on Machine Learning (ICML)}, 2020.

\bibitem{mamba}
A.~Gu and T.~Dao, ``Mamba: Linear-time sequence modeling with selective state
  spaces,'' {\em arXiv preprint arXiv:2312.00752}, 2023.

\bibitem{mamba2}
T.~Dao and A.~Gu, ``Transformers are {SSM}s: Generalized models and efficient
  algorithms through structured state space duality,'' in {\em International
  Conference on Machine Learning (ICML)}, 2024.

\bibitem{dosovitskiy2020vit}
A.~Dosovitskiy {\em et~al.}, ``An image is worth 16x16 words: Transformers for
  image recognition at scale,'' {\em arXiv preprint arXiv:2010.11929}, 2020.

\bibitem{chitta2022transfuser}
K.~Chitta, A.~Prakash, B.~Jaeger, Z.~Yu, K.~Renz, and A.~Geiger, ``Transfuser:
  Imitation with transformer-based sensor fusion for autonomous driving,'' {\em
  IEEE transactions on pattern analysis and machine intelligence}, vol.~45,
  no.~11, pp.~12878--12895, 2022.

\bibitem{touvron2021deit}
H.~Touvron, M.~Cord, M.~Douze, F.~Massa, A.~Sablayrolles, and H.~J{\'e}gou,
  ``Training data-efficient image transformers \{DeiT\},'' in {\em ICML}, 2021.

\bibitem{wang2021pvt}
W.~Wang {\em et~al.}, ``Pyramid vision transformer: A versatile backbone for
  dense prediction,'' in {\em ICCV}, 2021.

\bibitem{ranftl2021dpt}
R.~Ranftl, A.~Bochkovskiy, and V.~Koltun, ``Vision transformers for dense
  prediction,'' in {\em ICCV}, 2021.

\bibitem{wang2020linformer}
S.~Wang {\em et~al.}, ``Linformer: Self-attention with linear complexity,''
  {\em arXiv preprint arXiv:2006.04768}, 2020.

\bibitem{choromanski2020performer}
K.~Choromanski {\em et~al.}, ``Rethinking attention with performers,'' {\em
  arXiv preprint arXiv:2009.14794}, 2020.

\bibitem{kitaev2020reformer}
N.~Kitaev, {\L}.~Kaiser, and A.~Levskaya, ``Reformer: The efficient
  transformer,'' {\em arXiv preprint arXiv:2001.04451}, 2020.

\bibitem{beltagy2020longformer}
I.~Beltagy, M.~E. Peters, and A.~Cohan, ``Longformer: The long-document
  transformer,'' {\em arXiv preprint arXiv:2004.05150}, 2020.

\bibitem{zaheer2020bigbird}
M.~Zaheer {\em et~al.}, ``Big bird: Transformers for longer sequences,'' {\em
  arXiv preprint arXiv:2007.14062}, 2020.

\bibitem{child2019sparse}
R.~Child, S.~Gray, A.~Radford, and I.~Sutskever, ``Generating long sequences
  with sparse transformers,'' in {\em NeurIPS}, 2019.

\bibitem{lee2021fnet}
J.~Lee-Thorp, J.~Ainslie, I.~Eckstein, and S.~Onta{\~n}{\'o}n, ``Fnet: Mixing
  tokens with fourier transforms,'' {\em arXiv preprint arXiv:2105.03824},
  2021.

\bibitem{rao2021dynamicvit}
Y.~Rao {\em et~al.}, ``Dynamicvit: Efficient vision transformers with dynamic
  token sparsification,'' in {\em NeurIPS}, 2021.

\bibitem{liang2022evit}
Y.~Liang {\em et~al.}, ``Evit: Expediting vision transformers via token
  reorganization,'' in {\em ICLR}, 2022.

\bibitem{ryoo2021tokenlearner}
M.~S. Ryoo {\em et~al.}, ``Tokenlearner: What can 8 learned tokens do for
  images and videos?,'' in {\em NeurIPS}, 2021.

\bibitem{bolya2023tome}
D.~Bolya and J.~Hoffman, ``Token merging: Your vit but faster,'' {\em arXiv
  preprint arXiv:2303.17608}, 2023.

\bibitem{yang2022ats}
J.~Yang {\em et~al.}, ``Ats: Adaptive token sampling for efficient vision
  transformers,'' {\em arXiv preprint arXiv:2202.02973}, 2022.

\bibitem{samet1984quadtree}
H.~Samet, ``The quadtree and related hierarchical data structures,'' {\em ACM
  Computing Surveys}, vol.~16, no.~2, pp.~187--260, 1984.

\bibitem{tang2022quadtree}
S.~Tang, J.~Zhang, S.~Zhu, and P.~Tan, ``Quadtree attention for vision
  transformers,'' in {\em ICLR}, 2022.

\bibitem{sarlin2020superglue}
P.-E. Sarlin {\em et~al.}, ``Superglue: Learning feature matching with graph
  neural networks,'' in {\em CVPR}, 2020.

\bibitem{sun2021loftr}
J.~Sun {\em et~al.}, ``Loftr: Detector-free local feature matching with
  transformers,'' in {\em CVPR}, 2021.

\bibitem{dosovitskiy2017carla}
A.~Dosovitskiy, G.~Ros, F.~Codevilla, A.~Lopez, and V.~Koltun, ``Carla: An open
  urban driving simulator,'' in {\em Proceedings of the 1st Annual Conference
  on Robot Learning (CoRL)}, 2017.

\bibitem{dosovitskiy2020image}
A.~Dosovitskiy, L.~Beyer, A.~Kolesnikov, D.~Weissenborn, X.~Zhai,
  T.~Unterthiner, M.~Dehghani, M.~Minderer, G.~Heigold, S.~Gelly, J.~Uszkoreit,
  and N.~Houlsby, ``An image is worth 16x16 words: Transformers for image
  recognition at scale,'' {\em arXiv preprint arXiv:2010.11929}, 2020.

\bibitem{katharopoulos2020transformers}
A.~Katharopoulos, A.~Vyas, N.~Pappas, and F.~Fleuret, ``Transformers are rnns:
  Fast autoregressive transformers with linear attention,'' in {\em Proceedings
  of the 37th International Conference on Machine Learning (ICML)}, 2020.

\bibitem{gu2023mamba}
A.~Gu and T.~Dao, ``Mamba: Linear-time sequence modeling with selective state
  spaces,'' {\em arXiv preprint arXiv:2312.00752}, 2023.

\bibitem{dao2024transformersssm}
T.~Dao and A.~Gu, ``Transformers are ssms: Generalized models and efficient
  algorithms through structured state space duality,'' in {\em Proceedings of
  the 41st International Conference on Machine Learning (ICML)}, vol.~235 of
  {\em Proceedings of Machine Learning Research}, pp.~10041--10071, PMLR, 2024.

\bibitem{tolstikhin2021mlpmixer}
I.~O. Tolstikhin, N.~Houlsby, A.~Kolesnikov, L.~Beyer, X.~Zhai, T.~Unterthiner,
  J.~Yung, A.~Steiner, D.~Keysers, J.~Uszkoreit, M.~Lucic, and A.~Dosovitskiy,
  ``Mlp-mixer: An all-mlp architecture for vision,'' in {\em Advances in Neural
  Information Processing Systems (NeurIPS)}, 2021.

\bibitem{woo2023convnextv2codesigningscaling}
S.~Woo, S.~Debnath, R.~Hu, X.~Chen, Z.~Liu, I.~S. Kweon, and S.~Xie, ``Convnext
  v2: Co-designing and scaling convnets with masked autoencoders,'' 2023.

\end{thebibliography}

\end{document}